\documentclass[10pt,twocolumn,letterpaper]{article}

\usepackage{cvpr}              

\usepackage{multirow}
\usepackage{threeparttable}
\usepackage{array}
\usepackage{tikz}
\usetikzlibrary{arrows.meta,positioning,calc,fit,shapes.geometric}

\definecolor{cvprblue}{rgb}{0.21,0.49,0.74}
\usepackage[pagebackref,breaklinks,colorlinks,allcolors=cvprblue]{hyperref}

\def\paperID{*****} 
\def\confName{CVPR}
\def\confYear{2026}

\title{JFAA: Technical Report for the EPIC-KITCHENS-100 Action Anticipation Challenge at EgoVis 2026}

\author{
Qiaohui Chu$^{1\,2}$, Haoyu Zhang$^{1\,2}$, Yisen Feng$^{1}$, Meng Liu$^{3}$, Weili Guan$^{1}$, \\ Dongmei Jiang$^{2}$, Liqiang Nie$^{1}$\\
$^1$Harbin Institute of Technology (Shenzhen) \qquad  $^2$Pengcheng Laboratory    \\$^3$Shandong Jianzhu University\\
{\tt\small \{qiaohuichu8599, zhang.hy.2019, yisenfeng.hit, mengliu.sdu\}@gmail.com;} \\
{\tt\small \{honeyguan, nieliqiang\}@gmail.com; jiangdm@pcl.ac.cn}
}

\begin{document}
\maketitle
 \begin{abstract}
We propose \textbf{JFAA}, a \textbf{J}EPA-based \textbf{F}uture \textbf{A}ction \textbf{A}nticipation method for the EPIC-KITCHENS-100 (EK-100) Action Anticipation task. 
Inspired by the representation learning and future prediction ability of V-JEPA 2.1, JFAA uses a frozen encoder and predictor to extract observed context features and near-future latent tokens. 
A lightweight attentive probe is then trained to predict verb, noun, and action logits with separate task queries. 
To improve robustness, we further build a field-aware ensemble over selected epoch-level predictions, allowing each output field to benefit from its most reliable candidates. 
Experimental results on the official challenge server show that JFAA achieves first place in the EgoVis 2026 EK-100 Action Anticipation Challenge.
Our code will be released at \url{https://github.com/CorrineQiu/JFAA}.
\end{abstract}
    
\section{Introduction}
\label{sec:intro}

Egocentric video understanding has become an important research direction with the rapid growth of large-scale first-person video datasets. 
Unlike traditional third-person videos, egocentric videos are captured from the camera wearer's viewpoint. 
They naturally record the actor's attention, hand-object interactions, and subjective intentions.
This perspective is closely related to Embodied AI, where intelligent agents need to understand human behavior from an actor-centered viewpoint and provide timely assistance. 
However, egocentric videos are challenging to analyze. 
Frequent head and body movements introduce strong camera motion. 
Important objects may be partially occluded or visible only for a short time. 
The surrounding environment is also observed from a limited field of view. 
These properties have motivated a wide range of research problems, including action recognition, action anticipation, visual question answering, temporal localization, and object-centric reasoning. These related directions are complementary to action anticipation:
egocentric video question answering studies how to select visual evidence
for fine-grained context reasoning \cite{pmlr-v235-zhang24aj,zhang2025hcqa},
while multimodal and embodied-video studies model contextual semantics,
exocentric-to-egocentric knowledge transfer, and spatial structure for
understanding the camera wearer's intent and scene state
\cite{zhang2021multimodal,zhang2026exo2ego,zhang2026spatial}.

A representative benchmark in this area is the EK-100 Action Anticipation Challenge. 
It focuses on short-horizon egocentric action anticipation, where a model must predict the next action before the action becomes visible. 
Given a target action segment $A_i=[t_s^i,t_e^i]$, let $T_a$ denote the anticipation time and $T_o$ denote the observation time. 
The goal is to predict the verb, noun, and verb-noun action class of $A_i$ by observing only the preceding video interval:
\begin{equation}
\left[t_s^i-(T_a+T_o),\, t_s^i-T_a\right].
\label{eq:observation-window}
\end{equation}
For this challenge, $T_a$ is fixed to 1 second, and $T_o$ can be set by participants. 
No visual content after $t_s^i-T_a$ can be used. 
Thus, the model cannot rely on the target action itself. 
It must infer the future action from pre-action cues, such as hand motion, object context, scene state, and the camera wearer's intention. For future-action prediction, recent long-term action anticipation studies
emphasize hand-object interactions, verb-noun dependencies, and explicit
intention reasoning \cite{chu2026intention}. Challenge-winning Ego4D LTA
systems further show that foundation-model features, verb-noun modeling, and
language-model-based forecasting can improve long-horizon anticipation
\cite{chu2025technical}. Complementary episodic-memory localization work
indicates that early fusion of video cues is useful for temporal grounding in
untrimmed egocentric videos \cite{feng2025osgnet}.

\begin{figure*}[t]

  \centering

  \includegraphics[width=\linewidth]{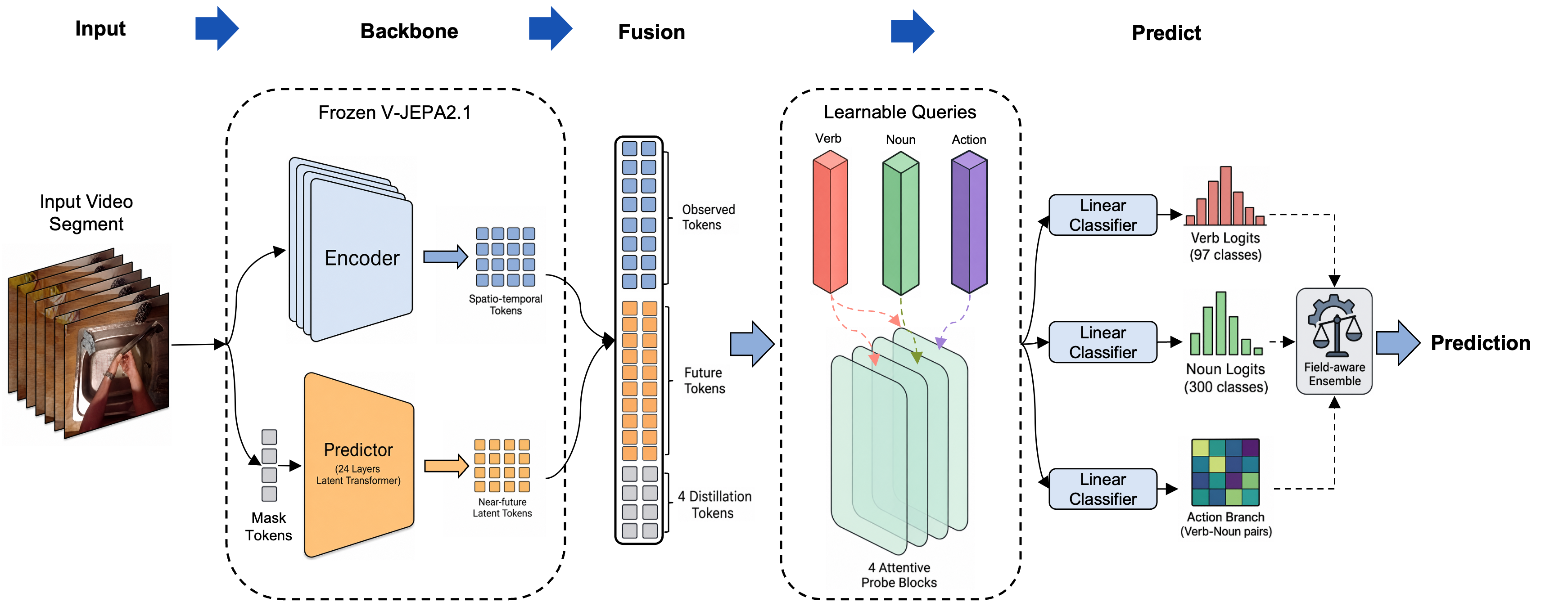}

  \caption{Overview of JFAA. JFAA extracts frozen V-JEPA 2.1 features from the observed clip, predicts near-future latent tokens, and uses attentive probes with field-aware ensemble inference.}

  \label{fig:method}

\end{figure*}

The challenge is conducted on EK-100, a large-scale unscripted egocentric video dataset collected from 45 kitchens in 4 cities. 
It contains about 100 hours of full HD video, 20M frames, 90K action segments, 20K unique narrations, 97 verb classes, and 300 noun classes~\cite{damen2022rescaling, damen2020epic}. 
The official evaluation uses Mean Top-5 Recall (MT5R) on the hidden test set. 
MT5R is computed by first measuring Top-5 Recall for each class and then averaging the results over all classes appearing in the test set. 
The metric is reported for three subsets: all test instances, instances from unseen participants, and tail classes. 
This protocol evaluates both general action anticipation ability and robustness under user shift and long-tailed class distributions.

To address this challenge, we propose \textbf{JFAA}, a \textbf{J}EPA based \textbf{F}uture \textbf{A}ction \textbf{A}nticipation method.
JFAA builds on V-JEPA 2.1~\cite{mur2026v} and uses a frozen ViT-G/384 encoder and predictor to obtain observed context features and near-future latent tokens. 
This design preserves the strong video representation ability of the foundation model while avoiding full backbone fine tuning. 
On top of these features, we train a lightweight attentive probe with separate queries for verb, noun, and action prediction. 
We further combine selected epoch predictions through a field-aware ensemble, so that the three output fields can use their most reliable candidates. 
Experimental results on the official challenge server show that JFAA achieves first place in the EgoVis 2026 EK-100 Action Anticipation Challenge.

\section{Method}
\label{sec:method}

Figure~\ref{fig:method} shows the overall pipeline of JFAA. 
The method first constructs pre-action clips from the observed video segment. 
It then uses a frozen V-JEPA 2.1 encoder-predictor backbone to extract observed-context features and near-future latent tokens. 
Finally, a lightweight attentive probe predicts verbs, nouns, and actions, and selected epoch-level predictions are combined by a field-aware ensemble.

\subsection{Clip Construction}

We use the official EK-100 frame folders and annotations. 
The probe is trained only on the official training split, while the official validation split is used for checkpoint evaluation, model selection, and qualitative analysis. 
Each action segment is converted into a pre-action clip. 
We sample 32 RGB frames at 8 FPS and resize or crop each frame to $384 \times 384$. 
During training, we randomly perturb the anticipation offset to expose the probe to different temporal contexts. 
During validation and testing, we follow the official setting and use a fixed 1.0-second anticipation gap. 
We apply standard video augmentations during training, including random resized cropping, RandAugment~\cite{cubuk2020randaugment}, horizontal flipping, and random erasing~\cite{zhong2020random}.

\subsection{Frozen V-JEPA 2.1 Backbone}

The visual backbone of JFAA is V-JEPA 2.1 ViT-G/384. 
We load the pretrained encoder and predictor and keep all backbone parameters frozen. 
Given an input clip, the encoder extracts spatiotemporal tokens from the observed frames. 
The predictor then estimates latent tokens for the near future. 
We concatenate the observed tokens from the encoder and the predicted tokens from the predictor. 
The resulting feature sequence contains both current visual evidence and future-predictive information. 
These features are then passed to the attentive probe for downstream prediction.

\subsection{Attentive Probe Classifier}

On top of the frozen features, we train a lightweight attentive classifier. 
The classifier contains four attentive probe blocks with 16 attention heads. 
Inside the attentive pooling module, we use three learnable queries for verb, noun, and action prediction. 
The pooled features are passed to three linear classifiers.

The verb and noun branches predict the corresponding label spaces, while the action branch predicts observed verb-noun action pairs. 
This design keeps the three prediction targets separate. 
Such separation is important for action anticipation, since verbs mainly depend on motion cues, while nouns rely more on object appearance and scene context. This design follows a general principle of decomposing prediction into
complementary semantic cues. Related attribute-guided learning work shows that
auxiliary semantic cues can help compensate for incomplete visual evidence in
partial visual recognition \cite{zhang2023attribute}. In JFAA, we instantiate
this general idea with task-specific verb, noun, and action queries rather than
explicit attribute annotations. To improve robustness, we train 20 probe heads with different learning-rate and weight-decay settings. 
This provides diverse candidates without fine-tuning the V-JEPA 2.1 backbone. 
We use sigmoid focal loss~\cite{lin2017focal} for verb, noun, and action prediction, since the action distribution is long-tailed. 
Training is performed with mixed precision.

\subsection{Prediction and Ensembling}

For each checkpoint, the model outputs scores for verb, noun, and action. Verb scores are expanded to the 97 official verb classes. Noun scores are expanded to the 300 official noun classes. Action scores are exported as the top 100 verb-noun pairs required by the challenge format. For each epoch, we evaluate all candidate probe heads on the validation split.
The best-performing head is selected and exported as one prediction candidate. The final submission is built from selected epoch predictions. We further apply a field-aware ensemble, where verb, noun, and action scores are selected and weighted separately. This design reduces the influence of a single checkpoint and improves test-time robustness.

\section{Experiments}
\label{sec:experiments}

\subsection{Experimental Protocol}

We follow the official EK-100 Action Anticipation protocol. 
The training split is used for model optimization, while the validation split is used for checkpoint evaluation and ensemble selection. 
No validation sample is used to optimize the model parameters. 
During validation and testing, the anticipation time is fixed to 1 second. 
The submitted file follows the official action anticipation format and contains scores for verb, noun, and action predictions. 
We report MT5R on the official validation split and final results on the hidden test set.

\subsection{Validation Results}

Table~\ref{tab:val_results} reports the validation MT5R of different epoch checkpoints. 
The results show that the best checkpoints for verb, noun, and action are not identical. 
Epoch 21 achieves the best verb MT5R, Epoch 26 achieves the best noun MT5R, and Epoch 22 achieves the best action MT5R. 
This observation supports our field-aware ensemble strategy, which selects and combines candidates separately for different output fields.

\begin{table}[t]
\centering
\caption{Validation Mean Top-5 Recall (MT5R, \%) on EK-100 action anticipation. Higher values indicate better performance.}
\label{tab:val_results}
\setlength{\tabcolsep}{8pt}
\begin{tabular}{lccc}
\toprule
Checkpoint & Verb $\uparrow$ & Noun $\uparrow$ & Action $\uparrow$ \\
\midrule
Epoch 18 & 60.9 & 59.3 & 39.1 \\
Epoch 19 & 62.3 & 58.8 & 38.8 \\
Epoch 20 & 63.9 & 60.3 & 39.2 \\
Epoch 21 & \textbf{65.1} & 60.3 & 38.9 \\
Epoch 22 & 63.6 & 61.1 & \textbf{39.6} \\
Epoch 23 & 64.0 & 60.8 & 39.5 \\
Epoch 24 & 64.2 & 61.2 & 39.5 \\
Epoch 25 & 65.0 & 60.1 & 39.5 \\
Epoch 26 & 64.4 & \textbf{61.8} & 39.1 \\
Epoch 27 & 64.3 & 60.5 & 39.2 \\
\bottomrule
\end{tabular}
\end{table}






\begin{table*}[htbp]
  \centering
  \caption{Official open-testing leaderboard on EK-100 action anticipation. Score denotes the overall action MT5R used for ranking. O, U, and T denote overall, unseen-participant, and tail-class subsets. V, N, and A denote verb, noun, and action. The corrine entry is our JFAA result.}
  \label{tab:leaderboard}
  \scriptsize
  \setlength{\tabcolsep}{2pt}
  \resizebox{\textwidth}{!}{%
  \begin{tabular}{r l r r r r r r r r r}
    \toprule
    Rank & Participant & Score & O-V & O-N & U-V & U-N & U-A & T-V & T-N & T-A \\
    \midrule
    1 & \textbf{corrine} & \textbf{27.95} & 49.22 & \textbf{52.02} & \textbf{47.83} & \textbf{57.89} & 30.90 & 42.86 & 41.53 & \textbf{21.65} \\
    2 & sunshinesky & 27.91 & 49.19 & 51.97 & \textbf{47.83} & 57.78 & \textbf{30.93} & 42.82 & 41.46 & 21.60 \\
    3 & abardes & 25.05 & 49.56 & 48.57 & 46.32 & 54.43 & 26.95 & 43.48 & 38.27 & 19.16 \\
    4 & InAViT IHPC-AISG-LAHA & 23.75 & 49.14 & 49.97 & 44.36 & 49.28 & 23.49 & 43.17 & 39.91 & 18.11 \\
    5 & OA-OAD & 22.96 & \textbf{51.57} & 45.97 & 44.02 & 39.04 & 16.71 & \textbf{48.50} & \textbf{41.55} & 19.24 \\
    6 & deleted\_user\_67855 & 9.32 & 21.15 & 35.71 & 18.79 & 36.62 & 9.56 & 13.54 & 27.33 & 6.51 \\
    7 & AIMS\_UNICAMP & 7.51 & 17.81 & 21.66 & 16.47 & 20.90 & 7.13 & 9.68 & 11.21 & 4.69 \\
    8 & itruonghai1 & 5.49 & 26.24 & 21.87 & 20.25 & 17.25 & 3.94 & 20.01 & 15.28 & 4.11 \\
    9 & itruonghai & 5.07 & 35.92 & 25.37 & 28.30 & 20.06 & 4.09 & 32.04 & 20.77 & 4.58 \\
    10 & eltoncn & 4.27 & 14.20 & 14.30 & 13.40 & 14.91 & 3.77 & 7.98 & 7.21 & 2.61 \\
    \bottomrule
  \end{tabular}
  }
\end{table*}

\subsection{Official Results}

Table~\ref{tab:leaderboard} reports the official open-testing leaderboard. 
JFAA ranks first in the challenge with an overall action MT5R of 27.95. 
It also shows strong performance on verb and noun prediction, as well as on unseen participants and tail classes. 
These results demonstrate the effectiveness of combining frozen V-JEPA 2.1 representations, near-future latent prediction, attentive probing, and field-aware ensemble inference.

\subsection{Case Study}

\begin{figure}[t]
  \centering
  \includegraphics[width=\columnwidth]{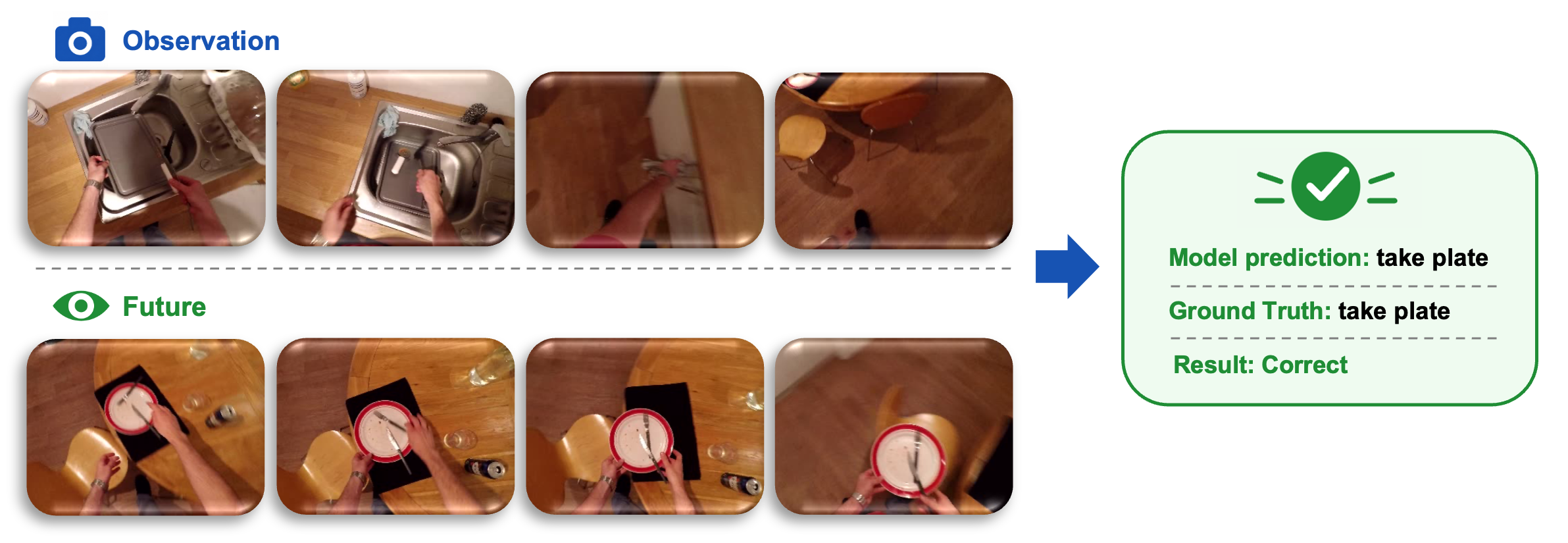}
  \caption{Successful case from the validation set. JFAA correctly predicts the future action \textit{take plate} from the observed pre-action frames.}
  \label{fig:case_success}
\end{figure}

\begin{figure}[t]
  \centering
  \includegraphics[width=\columnwidth]{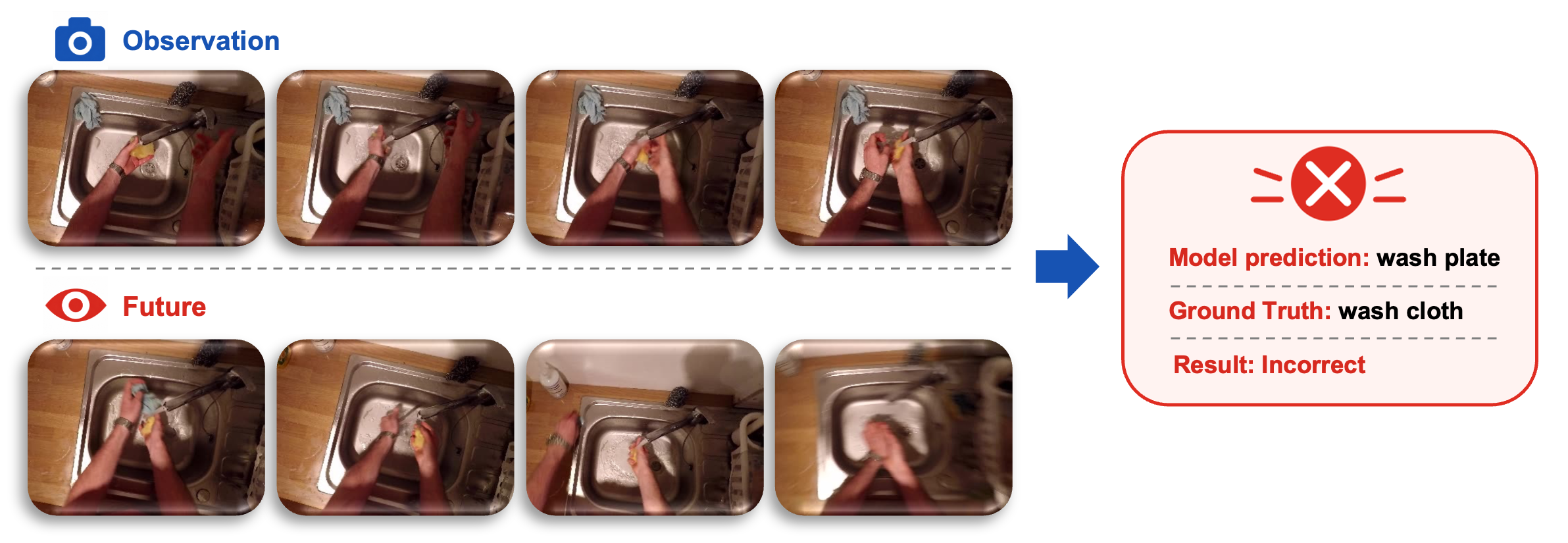}
  \caption{Failure case from the validation set. JFAA predicts \textit{wash plate} instead of the ground-truth action \textit{wash cloth}, showing a noun confusion in a visually similar sink context.}
  \label{fig:case_failure}
\end{figure}

Since the test labels are hidden, we conduct qualitative analysis on the validation set. 
Figures~\ref{fig:case_success} and~\ref{fig:case_failure} show a successful case and a failure case, respectively. 
In each case, the observed frames are sampled from the model input window, while the future frames are shown only for reference. 
These future frames are not observed by the model at inference time.

Figure~\ref{fig:case_success} shows a successful prediction. 
The visible plate and hand movement provide clear evidence for the upcoming action. 
JFAA correctly predicts the future action \textit{take plate}.

Figure~\ref{fig:case_failure} shows a failure case. 
The scene remains compatible with a washing action, but the objects around the sink are visually similar and contextually related. 
JFAA predicts \textit{wash plate} instead of the ground-truth action \textit{wash cloth}. 
This case suggests that noun prediction remains challenging when multiple candidate objects appear in similar contexts.

\section{Conclusion}
\label{sec:conclusion}

We introduce JFAA, a JEPA-based Future Action Anticipation method for the EK-100 Action Anticipation Challenge at EgoVis 2026. 
JFAA uses a frozen V-JEPA 2.1 ViT-G/384 encoder-predictor backbone to extract observed-context features and near-future latent tokens. 
A lightweight attentive probe is trained on top of these features to predict verbs, nouns, and actions with separate task queries. 
For the final submission, selected epoch-level predictions are further combined through a field-aware ensemble, allowing different output fields to benefit from their most reliable candidates.

Experimental results on the official challenge server show that JFAA achieves first place in the EgoVis 2026 EK-100 Action Anticipation Challenge and surpasses previous leading submissions. 
These results indicate that frozen future-predictive video representations, combined with lightweight probing and field-aware ensemble inference, provide an effective solution for EK-100 action anticipation.



{
    \small
    \bibliographystyle{ieeenat_fullname}
    \bibliography{main}
}


\end{document}